%% file: acl_latex.tex
\pdfoutput=1

\documentclass[11pt]{article}

\usepackage[final]{acl}

\usepackage{hyperref}

\usepackage{booktabs}  
\usepackage{graphicx}  
\usepackage{makecell}  
\usepackage{multirow}  

\usepackage{times}
\usepackage{latexsym}

\usepackage[T1]{fontenc}

\usepackage[utf8]{inputenc}

\usepackage{microtype}
\usepackage{adjustbox}

\usepackage{inconsolata}

\usepackage{graphicx}
\usepackage{subcaption}

\usepackage{algorithm}
\usepackage{amsmath}
\usepackage{algorithmic}
\usepackage{listings}
\title{\textsc{PerCoR}: Evaluating Commonsense Reasoning in Persian via Multiple-Choice Sentence Completion}

\author{
  Morteza Alikhani$^{\dagger}$ \quad
  Mohammadtaha Bagherifard$^{\dagger *}$ \quad
  Erfan Zinvandi \\
  \textbf{Mehran Sarmadi} \\
  MCINEXT \\
  \texttt{taha.bagheri98@gmail.com} \\
  \texttt{\{morteza.alikhani95, e.zeynvandi1376, mehran.sarmadi99\}@sharif.edu}
}

\begin{document}
\maketitle

\begingroup
\renewcommand\thefootnote{\fnsymbol{footnote}}
\footnotetext[2]{Equal contribution.}          
\footnotetext[1]{Corresponding author.}        
\endgroup
\begin{abstract}
We introduced \textbf{\textsc{PerCoR}}—\textbf{\emph{Per}}sian \textbf{\emph{Co}}mmonsense \textbf{\emph{R}}easoning—the first large-scale Persian benchmark for commonsense reasoning. \textsc{PerCoR} contains 106K multiple-choice sentence-completion problems drawn from more than forty news, cultural and other web sources. We adopt a linguistically grounded, conjunction-based segmentation strategy to generate coherent prefix–continuation pairs. To create challenging distractors, we propose \textbf{DRESS-AF}—\textbf{\emph{D}}istractor \textbf{\emph{R}}anking via \textbf{\emph{E}}mbedding \textbf{\emph{S}}imilarity \textbf{\emph{S}}coring and \textbf{\emph{A}}dversarial \textbf{\emph{F}}iltering—a generation-free adversarial filtering method that selects distractors from the pool of gold continuations while maximising model confusion.
Human annotators score 89\% on \textsc{PerCoR}, while \texttt{OpenAI-o3} achieves the highest performance at 92.18\%, followed closely by \texttt{Claude-Sonnet-3.7} (91.17\%). The strongest open-source model, \texttt{DeepSeek-R1}, reaches 82.51\%, underscoring both the dataset’s difficulty and the remaining performance gap in Persian commonsense reasoning. We further show that DRESS-AF transfers to the English HellaSwag benchmark, increasing its difficulty without hurting human solvability. The dataset is available at \url{https://huggingface.co/datasets/MCINext/PerCoR}.
\end{abstract}

\input{sections/introduction}

\input{sections/related_work}
\input{sections/method}

\input{sections/experiments}

\input{sections/conclusion}

\clearpage
\newpage
\section*{Limitations}
\input{sections/limitation}

\section*{Ethics}
\input{sections/ethics}

\bibliography{custom}

\appendix
\input{sections/appendix}

\end{document}

%% file: sections/introduction.tex
\section{Introduction}
\label{sec:intro}

Commonsense reasoning is a critical capability in natural language understanding, enabling models to draw inferences, disambiguate meaning, and interpret implicit knowledge. While large language models (LLMs) have shown remarkable progress across various tasks, their performance on commonsense reasoning—particularly in structured formats like multiple-choice sentence completion—remains limited \citep{sap-etal-2020-commonsense}. To benchmark and improve this ability, several datasets such as SWAG \citep{zellers-etal-2018-swag}, HellaSwag \citep{zellers-etal-2019-hellaswag}, and CommonsenseQA \citep{talmor-etal-2019-commonsenseqa} have been proposed. However, these benchmarks are overwhelmingly English-centric, leaving a significant gap in resources for evaluating and improving commonsense reasoning in low-resource languages.

\begin{figure}[t]
    \centering
    \includegraphics[width=\columnwidth]{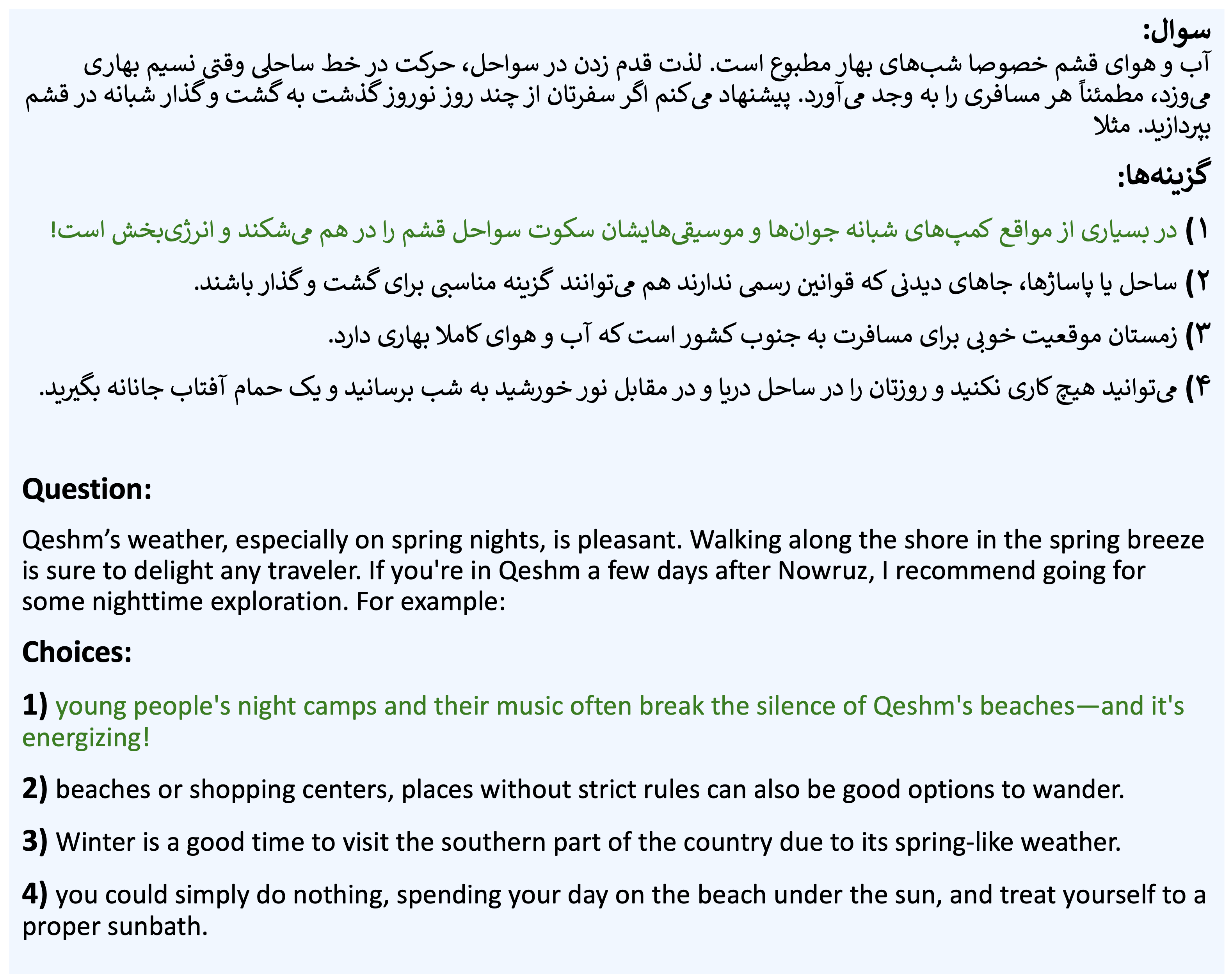}
    \caption{An example from the \textsc{PerCoR} dataset. The passage discusses the pleasant spring weather in Qeshm and recommends nighttime exploration. The correct answer (written in Green) refers to night camps and music breaking the beach's silence, while other options, though plausible in isolation, lack relevance to the immediate context.}
    \label{fig:percor-example}
\end{figure}

Despite recent progress in Persian NLP through resources such as \textsc{ParsiNLU} \citep{khashabi-etal-2021-parsinlu}, PersianQA \citep{PersianQA}, and \textsc{Pquad} \citep{Darvishi_2023}, Persian remains a low-resource language for high-level reasoning tasks, particularly commonsense inference. This leaves a significant gap in evaluating and advancing structured reasoning capabilities in this language. To address this limitation, we introduce \textbf{\textsc{PerCoR}}—the first large-scale Persian commonsense reasoning dataset in multiple-choice sentence–completion format. Constructed from over 40 diverse Persian websites, \textsc{PerCoR} captures a broad range of domains and linguistic styles. We formulate each instance as a sentence prefix followed by four completion candidates: one correct and three distractors. Instead of relying on simple rule-based methods for sentence segmentation, we generate sentence–completion pairs by splitting at conjunctions, promoting natural flow and semantic coherence. This connective-driven formulation is related to prior work that uses discourse markers for inference and sentence completion~\citep{bhargava-ng-2022-discosense}, but \textsc{PerCoR} applies this strategy at scale to Persian and across diverse, non-narrative domains. Unlike oversimplified strategies such as the one employed in SWAG~\citep{zellers-etal-2018-swag}, which relies on temporally grounded data like video captions, our conjunction-based approach is applicable to a wide range of textual sources. This enables broader domain coverage and greater variability in sample length, enhancing both the diversity and richness of the dataset.

\begin{figure*}[h]
    \centering
    \includegraphics[width=\textwidth]{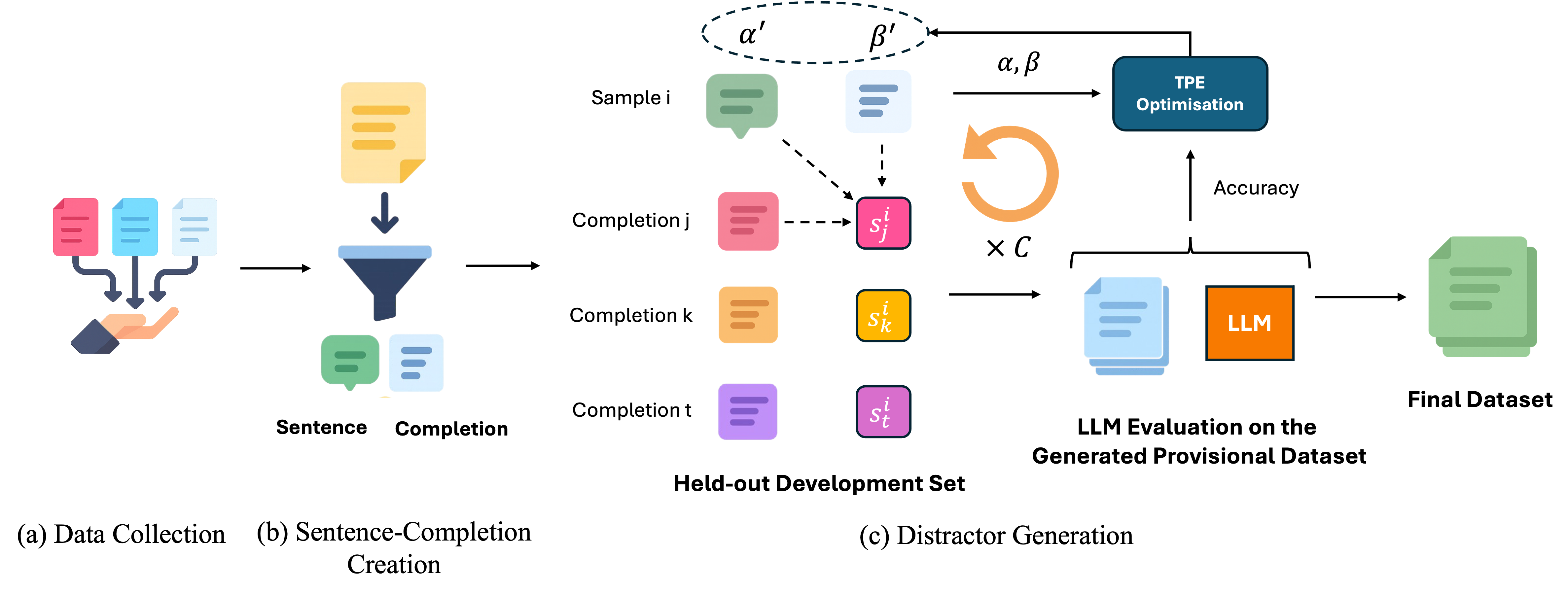}

    \caption{Overview of our dataset construction and distractor generation pipeline. The process consists of: (a) collecting diverse Persian text data, (b) creating and filtering sentence-completion pairs, and (c) generating challenging multiple-choice distractors using \textbf{DRESS-AF}.}

    \label{fig:dataset-pipeline}
\end{figure*}

We further propose a novel distractor selection strategy, DRESS-AF, which is a combination of Adversarial Filtering (AF) \citep{zellers-etal-2019-hellaswag} and Embedding-based ranking \citep{liang-etal-2018-distractor,chiang-etal-2022-cdgp} methods. DRESS-AF avoids LLM-based generations—thus sidestepping associated biases—and instead ranks completions using embedding-based similarity metrics. These scores are adversarially tuned to maximise model confusion using Bayesian optimisation over a development set, yielding difficult yet human-solvable distractors.

\textsc{PerCoR} specifically targets discourse-level and contextual commonsense reasoning. To select the correct continuation, a model must infer the semantic relation introduced by the conjunction (e.g., causal, contrastive, elaborative) and identify which candidate maintains coherence with the preceding context. Distractors are intentionally plausible in isolation but violate these discourse constraints, requiring models to integrate implicit world knowledge with local coherence cues rather than relying on shallow lexical matching.

An example is shown in Figure~\ref{fig:percor-example} illustrating a key aspect of \textsc{PerCoR} dataset—candidates are intentionally context-sensitive. While all options may appear semantically valid in isolation, only one logically follows from the passage. In this case, the mention of “nighttime exploration” cues the correct choice, requiring the model to interpret implicit temporal references to succeed.

In summary, our key contributions are as follows: \textbf{(1)} we introduce \textbf{\textsc{PerCoR}}, the first large-scale Persian commonsense reasoning dataset in a multiple-choice sentence–completion format, spanning diverse domains and linguistic styles; \textbf{(2)} we propose a conjunction-based extraction method that enables natural and semantically coherent sample generation from non-temporal texts; \textbf{(3)} we present \textbf{DRESS-AF}, a language-agnostic, embedding-based distractor generation approach that incorporates adversarial filtering to produce challenging yet human-solvable distractors—without relying on generative models; and \textbf{(4)} we benchmark a broad set of state-of-the-art open- and closed-source LLMs on PerCoR, establishing strong empirical baselines for future work.

%% file: sections/related_work.tex
\section{Related Work}

\paragraph{Commonsense Reasoning Datasets.}
Numerous English benchmarks have been introduced to evaluate commonsense reasoning in multiple formats. \textsc{SWAG} \citep{zellers-etal-2018-swag} and \textsc{HellaSwag} \citep{zellers-etal-2019-hellaswag} pose multiple-choice sentence completion tasks based on narrative or descriptive contexts. HellaSwag, in particular, uses adversarial filtering to create distractors that are challenging for language models but easily solvable by humans. Cloze-style narrative continuation tasks such as the \textsc{Story Cloze Test}~\citep{mostafazadeh-etal-2016-corpus} also involve selecting a plausible continuation, though they focus on narrative coherence rather than connective-driven discourse inference.
Other benchmarks such as \textsc{WinoGrande} \citep{SakaguchiWinnogrande}, \textsc{CommonsenseQA} \citep{talmor-etal-2019-commonsenseqa}, \textsc{OpenBookQA} \citep{mihaylov-etal-2018-suit}, \textsc{PIQA} \citep{Bisk2019PIQARA}, \textsc{Cosmos} \citep{huang-etal-2019-cosmos}, and \textsc{Social IQa} \citep{sap-etal-2019-social} cover a variety of commonsense dimensions, including physical reasoning, social dynamics, and multi-hop inference. More recent efforts include \textsc{GLUCOSE} \citep{mostafazadeh-etal-2020-glucose}, a dataset of causal explanations in short narratives, annotated across ten dimensions of inferential knowledge; \textsc{COM2SENSE} \citep{singh-etal-2021-com2sense}, which evaluates a model’s ability to discriminate between true and false commonsense statements in complementary pairs; and \textsc{CommonsenseQA 2.0} \citep{talmor2021commonsenseqa}, an adversarially curated yes/no question dataset designed to be difficult for large language models while remaining easy for humans. Despite substantial progress, these benchmarks are primarily designed for English, leaving a gap in resources for other languages.

\paragraph{Distractor Generation Techniques.}
Creating high-quality distractor candidates is crucial for constructing reliable multiple-choice datasets \citep{alhazmi2024distractorgenerationmultiplechoicetasks}. \textit{Adversarial filtering} (AF), used in \textsc{SWAG}, \textsc{HellaSwag}, and \textsc{WinoGrande}, iteratively removes easy distractors using a discriminator model, resulting in semantically challenging options. Alternatively, \textit{retrieval-based methods} select distractors from external corpora or knowledge graphs, ensuring topical relevance \citep{ren21-knowledge-driven}. Recent work extends this by incorporating topic models to filter noisy candidates from knowledge graphs like Probase \citep{ren2021knowledge}. \textit{Embedding-based ranking} selects distractors based on similarity in embedding space \citep{liang-etal-2018-distractor, chiang-etal-2022-cdgp}, while \textit{retrieval-augmented generation} leverages retrieved passages and knowledge triplets to guide large language models in producing diverse distractors \citep{chen2023retrieval}. Our proposed method, DRESS-AF, combines these principles: we rank gold completions using embedding-based similarity scoring and adversarially optimise parameters to select distractors that maximise model confusion.

\paragraph{Persian NLP Resources.}
Recent years have seen a growing body of work on Persian NLP, but most resources target core tasks such as machine translation, sentiment analysis, and reading comprehension. \textsc{ParsiNLU} \citep{khashabi-etal-2021-parsinlu} includes benchmarks for Persian NLI, QA, and sentiment classification. \textsc{FarsTail} \citep{Amirkhani_2023} is a natural language inference dataset, while \textsc{PQuAD} \citep{Darvishi_2023} provides large-scale reading comprehension benchmarks in the SQuAD format. For more open-ended reasoning, \textsc{PerCQA} \citep{jamali-etal-2022-percqa} is a community QA dataset compiled from Persian web forums, consisting of 989 real-world questions and over 21k answers, designed for tasks like answer selection and ranking. Although these resources enable evaluation of Persian understanding and reasoning, they do not address commonsense reasoning specifically. To the best of our knowledge, our work presents the first large-scale Persian commonsense reasoning dataset, addressing a significant gap in low-resource language evaluation.

%% file: sections/method.tex
\section{The \textsc{PerCoR} Dataset}
\label{sec:method}
We adopt a three-stage pipeline to create the \textsc{PerCoR} dataset: (1)~\textbf{Data Collection}, in which raw text segments are gathered from diverse sources; (2)~\textbf{Sentence–Completion Creation}, where sentence–completion pairs are generated using our novel conjunction-based method; and (3)~\textbf{Distractor Generation}, where we apply our proposed DRESS-AF algorithm to select challenging distractor candidates for each instance. An overview of this pipeline is illustrated in Figure~\ref{fig:dataset-pipeline}.

\subsection{Data Collection}
To construct our dataset, we begin by collecting a diverse set of paragraphs spanning a broad range of topics, ensuring that meaningful sentence–completion pairs can be extracted for multiple-choice commonsense evaluation. For this purpose, we leverage the Corpesia corpus \citep{sarmadi2025hakimfarsitextembedding}, a large-scale resource built by crawling the main content (excluding advertisements and irrelevant sections) from a wide variety of Persian websites. The raw data in Corpersia is cleaned through rule-based filtering to remove boilerplate artifacts—such as author names, timestamps, and footers—while preserving the original paragraph structure and maintaining document-level segmentation, including title identification.

To generate sentence–completion pairs, we select a subset of websites from Corpersia that cover a broad spectrum of topics (detailed in Section~\ref{sec:experiments-stats}). We discard any paragraph with fewer than 50 characters to ensure the textual quality and context richness. Finally, we sample up to 200{,}000 paragraphs from each selected website to be used as the source for extracting sentence–completion pairs.

\subsection{Sentence–Completion Creation}
Rather than relying on conventional techniques such as those employed in SWAG \citep{zellers-etal-2018-swag}, which depend on temporally coherent data (e.g., video captions), we adopt a linguistically grounded strategy based on conjunctions to extract sentence–completion pairs from static text. Specifically, we begin by curating a list of 49 high-frequency conjunctions in Persian. To ensure consistency and reduce sparsity, we remove conjunctions that appear fewer than 500 times in the corpus. Importantly, all excluded items have semantically equivalent counterparts in the retained set, preserving the expressivity of the conjunction space. The final list of conjunctions, along with their English translations, is presented in Figure~\ref{fig:conjunction_list}. To maintain balanced representation across conjunctions and avoid dominance by high-frequency items, we sample up to 4,000 instances per connective. If a conjunction occurs fewer than 4,000 times, we include every instance. For semantically ambiguous conjunctions—those that may not always function as true connectives in contexts—we increase our oversampling multiplier so that the filtered data retains a sufficient number of valid usages.

To ensure that the sentence–completion split occurs at an informative and coherent boundary, we define a valid character span within which conjunctions are considered—ranging from a minimum of 50 to a maximum of 250 characters from the start of the paragraph. The lower bound ensures that the prefix contains sufficient context for prediction, while the upper bound prevents overly long or semantically overloaded prefixes. Once a valid conjunction is found within this range, we check the character length of the clause following it. If the length is below a threshold of 150 characters, the paragraph is split at that conjunction to form a sentence–completion pair. Otherwise, the search continues with other conjunctions in the span. This ensures that the completion segment remains concise and focused.

To further validate the quality of extracted pairs , we perform a lightweight filtering step using the \texttt{GPT-4o-mini} model. Specifically, the model is used for two binary classification checks: (1) verifying that the identified conjunction functions as a true discourse connective (since some Persian conjunctions may be contextually ambiguous), and (2) ensuring that the completion segment is a syntactically and semantically complete sentence. Since the model is only used for verification, not generation, it does not introduce generation-related biases into the data. Additional details regarding this filtering process are provided in Appendix~\ref{sec:appendix-completion}.

\subsection{Distractor Generation}

To avoid introducing any biases associated with language model generations, we select distractor options from the set of gold completions belonging to other samples, rather than generating them via an LLM. Let $\mathbf{x}_i$ and $\mathbf{y}_i$ be the embedding of the sentence and completion, respectively. We define a score \(s^i_j\), representing the suitability of \(\text{completion}_j\) as a candidate option for \(\text{sentence}_i\), as follows:
\begin{equation}
\begin{aligned}
s^i_j =\;& \alpha \cos(\mathbf{x}_i, \mathbf{y}_j)
    + \beta \cos(\mathbf{y}_i, \mathbf{y}_j) \\
      &+ (1 - \alpha - \beta) \cos(\mathbf{z}_i, \mathbf{y}_j),
\end{aligned}
\end{equation}
where \(\alpha, \beta \in [0,1]\) are tunable coefficients that balance the contributions of each similarity term, \(\cos(\cdot,\cdot)\) denotes cosine similarity, and $\mathbf{z}_i$ refers to the embedding of concatenation of the sentence and its gold completion. Using a held-out development set, we compute \(s^i_j\) for each sample pair \(i\) within the development set and all candidates \(j\) within the whole data (not only in the development set). Based on these scores, we sort the candidates in descending order, exclude the gold completion \(\mathbf{y}_i\), and uniformly sample three distractors from the next \(k\)-best candidates. This process yields a 4-way multiple-choice instance for each sentence in the held-out set, constructed dynamically according to the current values of \(\alpha\) and \(\beta\).

We optimize \(\alpha\) and \(\beta\) via adversarial filtering: for a given \((\alpha,\beta)\), we build the provisional dataset from the held-out development set, measure the accuracy of an LLM on it, and use that accuracy as the objective in a Tree-structured Parzen Estimator (TPE) Bayesian optimization over \(c\) trials. Although the search space is low-dimensional, TPE is known for its strong empirical performance and sample efficiency in hyperparameter tuning tasks, and has been widely adopted in AutoML and deep learning optimisation pipelines \citep{ watanabe2023treestructuredparzenestimatorunderstanding}.\footnote{We employed the implementation of TPE in the \texttt{Optuna} library~\citep{akiba2019optuna}.}. The optimal \((\alpha^*,\beta^*)\) are then used to generate our final dataset. To construct the \textsc{PerCor} dataset, we set \(c=30\) and \(k=20\). We employed the \textsc{Hakim} embedding model \citep{sarmadi2025hakimfarsitextembedding}, as it demonstrated the best performance on \textsc{FaMTEB} \citep{zinvandi2025famtebmassivetextembedding}, a comprehensive benchmark for Persian text embeddings.

We refer to this method as \textbf{DRESS-AF} (\textit{Distractor Ranking via Embedding Similarity Scoring and Adversarial Filtering}). DRESS-AF constructs multiple-choice questions by scoring all candidate completions using the embedding-based metric defined above, and then adversarially optimising the scoring parameters to select the most challenging distractors. Importantly, the adversarial nature of DRESS-AF ensures that the selected distractors increase question difficulty—but it does not guarantee overall dataset quality or standardness. In practice, two hyperparameters play a key role in adjusting the difficulty: \(c\), the number of optimisation trials, and \(k\), the number of top-ranked distractor candidates (after excluding the gold completion) from which three distractors are randomly sampled. While DRESS-AF aims to generate difficult examples for evaluation, human oversight may still be required to discard samples that are excessively ambiguous or unsolvable, ensuring that the final dataset remains reliable and informative.

We hypothesise that the set of gold completions across all samples is sufficiently diverse to serve as a reliable pool of distractor candidates. This assumption enables us to avoid synthetic generation altogether and sidestep potential biases introduced by LLM outputs. In Section~\ref{sec:experiments}, we empirically validate this hypothesis by showing that several strong LLMs consistently achieve below 80\% accuracy on our dataset. This confirms the overall challenge posed by the distractors selected via DRESS-AF. Furthermore, evaluations conducted by human annotators on a subset of the data yield accuracies around 90\%, providing additional evidence that the questions are both plausible and solvable, albeit non-trivial.

%% file: sections/experiments.tex
\section{Experiments}
\label{sec:experiments}
We structure our experiments in three phases: first, we analyse the \textsc{PerCoR} dataset by examining token-length distributions and covered topics and domain; second, we evaluate DRESS-AF's ability to craft challenging distractors for the sentence-completion pairs in the HellaSwag dataset, demonstrating the generality of our method in generating strong distractors without relying on generative models, and also its applicability beyond Persian; third, we benchmark 32 large language models on the dataset in a zero-shot setting to gauge their out-of-the-box performance. Further experiments regarding the effect of input length and also few-shot evaluation are provided in Appendix~\ref{sec:appendix-model-len},~\ref{sec:appendix-fewshot}.

\subsection{Dataset Statistics}
\label{sec:experiments-stats}
The dataset is divided into three splits: training (86,217 samples), validation (10,000 samples), and test (10,000 samples). Each sample consists of an uncompleted text and four candidate completions. The average sentence length is 129.23 characters and 41.78 tokens, while the average completion length is 93.24 characters and 30.08 tokens. Completion statistics are computed by first averaging the length (in characters and tokens) across the four candidates within each sample, and then taking the mean over all samples. Token lengths are calculated using the \textit{GPT-4o-mini} tokeniser via the \texttt{tiktoken} library \citep{tiktoken}.

To ensure linguistic and topical diversity in our dataset, we collected raw Persian text data from over 40 distinct websites spanning a broad range of domains. These include news and current affairs (e.g., \href{https://www.isna.ir/}{ISNA}, \href{https://www.khabaronline.ir/}{KhabarOnline}, \href{https://www.yjc.ir/}{YJC}), technology and digital media (e.g., \href{https://digiato.com/}{Digiato}, \href{https://www.zoomit.ir/}{Zoomit}), religion and culture (e.g., \href{https://hawzah.net/}{Hawzah}, \href{https://fa.wikishia.net/}{WikiShia}, \href{https://wiki.ahlolbait.com/}{Wiki Ahlolbait}), lifestyle and health (e.g., \href{https://www.ninisite.com/}{NiniSite}, \href{https://doctoreto.com/}{Doctoreto}, \href{https://namnak.com/}{Namnak}), economy and business (e.g., \href{https://www.eghtesadonline.com/}{EqtesadOnline}, \href{https://www.ecoiran.com/}{Ecoiran}, \href{https://www.digikala.com/mag/}{Digikala Mag}), travel and leisure (e.g., \href{https://hamgardi.com/}{Hamgardi}, \href{https://www.alibaba.ir/}{Alibaba}), education and self-improvement (e.g., \href{https://fidibo.com/}{Fidibo}, \href{https://taaghche.com/}{Taaghche}, \href{https://motamem.org/}{Motamem}), and sports and entertainment (e.g., \href{https://www.varzesh3.com/}{Varzesh3}, \href{https://www.vipofilm.com/}{VIPofilm}). In addition, user-generated content platforms like \href{https://virgool.io/}{Virgool} contribute informal and diverse writing styles. This domain variety enables broad coverage of content structures, writing registers, and topics, making the dataset a representative resource for real-world commonsense reasoning in Persian.

\subsection{Effectiveness of DRESS-AF in Distractor Generation}
\label{sec:effect-dress}
\subsubsection{PerCoR Dataset}
To evaluate the effectiveness of our proposed distractor generation method during the construction of the dataset, we track the performance of the \texttt{GPT-4o-mini} model on the provisional datasets constructed during the optimization process. Specifically, we run \(c = 30\) trials, where in each trial we use a different pair of \((\alpha, \beta)\) coefficients to generate distractors based on the scoring function defined in Section~\ref{sec:method}. Our goal is to adversarially reduce the model's accuracy—i.e., to identify distractor settings that make the multiple-choice task more challenging. Among the \(c\) trials, we select the \((\alpha^*, \beta^*)\) pair corresponding to the lowest model accuracy for use in the final dataset construction.

\begin{figure}[t]
    \centering
    \includegraphics[width=\columnwidth]{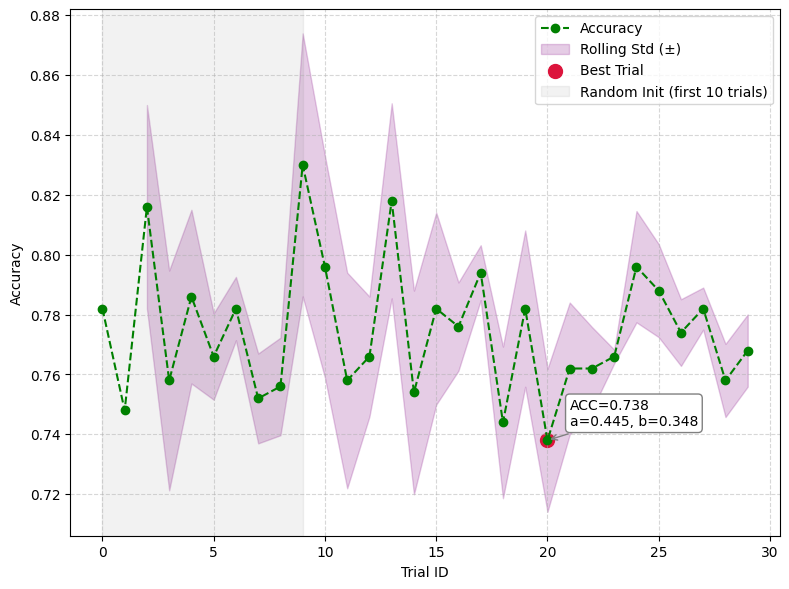}
    \caption{Accuracy of GPT-4o-mini on the provisional dataset, during the construction of the PerCoR dataset. DRESS-AF tries to find the best coefficients within 30 trials. The first 10 trials use random sampling, followed by TPE-based search. The lowest accuracy (trial 20) corresponds to the selected distractor configuration.}
    \label{fig:trial-accuracy-persian}
\end{figure}

Figure~\ref{fig:trial-accuracy-persian} shows the model's accuracy across the 30 trials. For the first 10 trials, we use random initialization to encourage exploration; from trial 11 onward, we apply the Tree-structured Parzen Estimator (TPE) algorithm for guided search. We plot the accuracy along with a rolling standard deviation (window size = 3) to visualize exploration dynamics. As seen, the variance is initially high due to random sampling, then decreases as the optimization converges. The lowest observed accuracy occurs at trial 20, indicating the most adversarial configuration found by DRESS-AF.

\subsubsection{HellaSwag Dataset}
To further demonstrate the effectiveness and generality of DRESS-AF in generating challenging distractor candidates without introducing generation-induced biases from LLMs, we apply the method to a non-Persian benchmark: the HellaSwag dataset \citep{zellers-etal-2019-hellaswag}. Specifically, we take the validation split of the HellaSwag dataset, then use its sentence--completion pairs (i.e., the context and gold ending) as inputs to DRESS-AF, showcasing the method’s language-agnostic applicability. 

To evaluate the extent to which DRESS-AF allows control over distractor difficulty, we construct two new variants of HellaSwag. In the first (harder) version, for each sample, we randomly sample three distractors from the top 10 highest-scoring candidates based on the embedding similarity score (excluding the gold completion). In the second (easier) version, we exclude the top 3 candidates (and the gold completion if it is not among them), then sample three distractors from the next top 20. For both versions, we run \(c = 30\) optimisation trials to find the best \((\alpha, \beta)\) parameters via the DRESS-AF procedure. For both variations, we employed Jina-v3 \citep{sturua2024jinaembeddingsv3multilingualembeddingstask} as the embedding model.

\begin{figure}[t]
    \centering
    \includegraphics[width=\columnwidth]{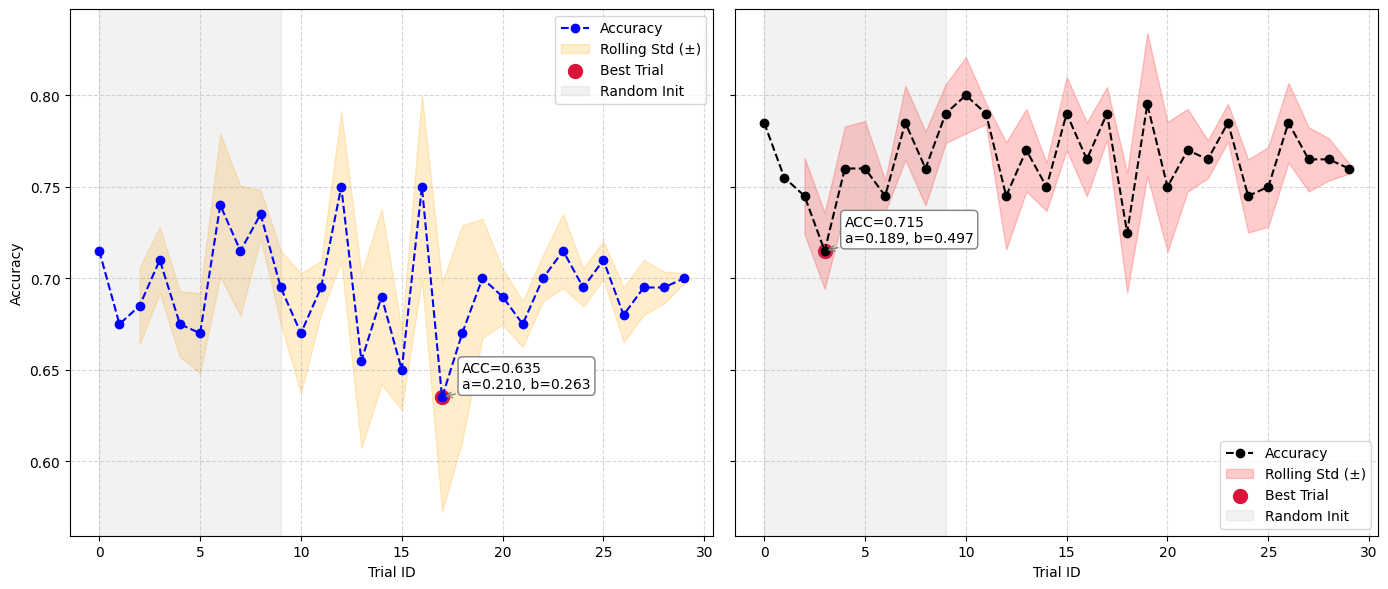}
    \caption{Accuracy of GPT-4o-mini on the provisional dataset across 30 trials during DRESS-AF optimisation on sentence--completion pairs from HellaSwag. The left plot corresponds to the harder version, with distractors sampled from the top 10 candidates. The right plot corresponds to the easier version, where the top 3 candidates are excluded and distractors are sampled from the next top 20.}
    \label{fig:trial-accuracy-hswag-dif}
\end{figure}

Figure~\ref{fig:trial-accuracy-hswag-dif} shows the accuracy of \texttt{GPT-4o-mini} on provisional datasets over the 30 trials during the tuning process of \((\alpha, \beta)\). The observed trend resembles the Persian setup in Figure~\ref{fig:trial-accuracy-persian}: during the initial 10--15 randomly sampled trials, variance is high due to exploration; afterward, performance stabilises as TPE converges. Using the best-found \((\alpha, \beta)\), we finalise the two dataset variants. We then evaluate both closed-source (\texttt{GPT-4o-mini}) and open-source (\texttt{Gemma-3-27B-it}) models on these variants, as well as on the original HellaSwag, to assess how distractor difficulty affects performance.

\begin{figure}[t]
    \centering
    \includegraphics[width=\columnwidth]{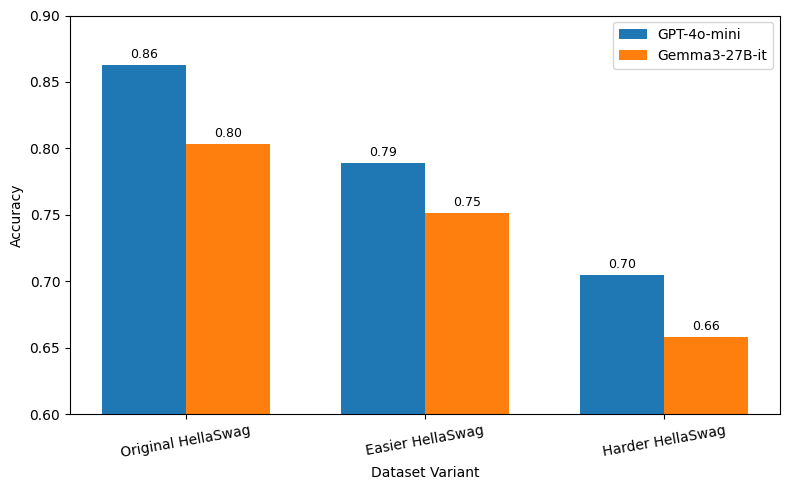}
    \caption{Accuracy of \texttt{GPT-4o-mini} and \texttt{Gemma3-27B-it}, representing closed- and open-source models respectively, across three HellaSwag variants. DRESS-AF was used to generate distractors for the easier and harder variants.}
    \label{fig:hswag-acc}
\end{figure}

Figure~\ref{fig:hswag-acc} presents model performance across three versions of the HellaSwag dataset: the original, an easier variant, and a harder one—both constructed using DRESS-AF. As expected, accuracy decreases as the distractor difficulty increases, demonstrating the method's effectiveness in producing more challenging distractors. Notably, both \texttt{GPT-4o-mini} and \texttt{Gemma3-27B-it} exhibit the lowest accuracy on the harder variant, indicating that DRESS-AF successfully identifies distractors that are more confounding for models.

The performance difference between the easier and harder versions can be attributed to the distractor sampling strategy. In the easier variant, we exclude the top three most confounding candidates and then randomly select from the next top 20. This design favours broader semantic differences between gold and distractor completions. In contrast, for the harder variant, we randomly sample distractors from the top 10 candidates, making the distractors more semantically similar to the correct answer and hence more difficult.

While increasing distractor difficulty is desirable, it is equally important that the dataset remain answerable and free of generation-induced artifacts. To assess this, we conduct a human evaluation on a 200-sample subset of each variant. On the original HellaSwag, human accuracy is 90\%. This remains nearly unchanged on the easier variant (89.5\%) and decreases only modestly on the harder one (83\%). In contrast, model accuracy drops much more sharply: for \texttt{Gemma-3-27B-it} and \texttt{GPT-4o-mini}, performance decreases by 5--7 points on the easier set and by 14--16 points on the harder set (Figure~\ref{fig:hswag-acc}). This asymmetry indicates that models were partially exploiting stylistic or fluency artifacts present in the original LM-generated distractors—artifacts that humans did not rely on. By selecting distractors from real human-written continuations, DRESS-AF reduces such generation-related biases and forces models to rely more on contextual and commonsense reasoning. At the same time, the modest change in human accuracy confirms that the resulting distractors remain human-solvable. In sum, DRESS-AF increases difficulty while maintaining dataset integrity and provides a more robust evaluation signal than generation-based AF methods.

\subsection{Model Results on \textsc{PerCoR}}\label{sec:results-percor}

To assess the \emph{out-of-the-box} commonsense abilities of modern LLMs in Persian, we evaluated 12 closed-source and 20 open-source models in a \textbf{zero-shot, multiple-choice} setting. Each model was prompted in Persian to return \emph{only} the index of the correct option. We report: (1) \textbf{Strict Accuracy}—exact match on the raw output, and (2) \textbf{Post-Processed Accuracy}—after applying a simple regex to extract the final digit \texttt{1–4}, recovering correct answers when extra justification is included. The model results appear in Table~\ref{tab:percor-eval}.

Overall, closed‐source models dominate: \texttt{OpenAI-o3} \citep{openai-o3} tops the leaderboard at \textbf{92.18\,\%}, followed by \texttt{Claude-3.7-Sonnet} \citep{anthropic-claude3.7-sonnet} (91.17\,\%) and \texttt{GPT‑4.1} \citep{openai-gpt4.1} (88.39\,\%); the best open‐source checkpoint, \texttt{DeepSeek‑R1} \citep{deepseekai2025deepseekr1incentivizingreasoningcapability}, reaches \textbf{82.51\,\%}, trimming the gap to roughly 10\%, while most open‐source peers fall between 60\,\% and 80\,\%. Human majority‐vote accuracy on \textsc{PerCoR} is 89\,\% (details in Appendix~\ref{sec:appendix-humaneval}), so only \texttt{o3} and \texttt{Sonnet} currently exceed non‐expert annotators performance. Despite strong aggregate performance, top-performing models still exhibit occasional failures on nuanced reasoning cases—several examples are provided in Appendix.

Formatting sensitivity is revealed by the gaps between Strict and Post‐Processed accuracy: e.g., \texttt{GPT‑4o} \citep{openai-gpt4o} from 78.32\% to 86.65\% (+8.3\%), \texttt{LLaMA‑3.3-70B} \citep{grattafiori2024llama3herdmodels} from 11.23\% to 79.56\% (+68.3\%), \texttt{Aya-Expanse-32B} \citep{dang2024ayaexpansecombiningresearch} from 5.85\% to 63.27\% (+57.4\%), and \texttt{DeepSeek‑V3} \citep{deepseekai2025deepseekv3technicalreport} from 51.15\% to 82.41\% (+31.3\%). Large difference indicates that the model often embeds the correct answer in extra prose; shallow post‐processing recovers more than 60\% of hidden accuracy for some models. In contrast, other models show consistent and similar accuracies, indicating strong adherence to the required output format.

Within individual open-source families, accuracy generally scales with parameter count: the \texttt{Gemma3} \citep{gemmateam2025gemma3technicalreport} series improves from 26\% (1B) to 76\% (27B), \texttt{Qwen‑3} \citep{yang2025qwen3technicalreport} from 50\% (4B) to 76.5\% (32B), while \texttt{Mistral} \citep{jiang2023mistral7b,mistral-small-31} lags (7B instruct: 30\%; 24B “Small-3.1”: 69\%). \texttt{Command A} \citep{cohere2025commandaenterprisereadylarge} outperforms its predecessor \texttt{Command R} \citep{cohere-command-r} (79.8\,\% vs.\ 60.0\,\%), likely due to its significantly larger parameter count and improved multilingual alignment—especially in Persian. The \texttt{LLaMA‑3.2} instruction variants (1B/3B) underperform (<25\%), yet the 70B variant, after post‐processing, rivals \texttt{Gemma3‑27B}. These trends confirm that parameter count alone is insufficient; alignment strategy and prompt‐format robustness are equally critical on PerCoR.

Closed‐source diversity also emerges: \texttt{o3}~{>}\,~\texttt{GPT‑4.1}~{>}\,~\texttt{GPT‑4o} suggests benefits from more advanced architecture and reasoning abilities. While \texttt{OpenAI-o4-mini} belongs to the same “o-series” family, it underperforms \texttt{o3} by a notable margin (85.5\,\% vs.\ 92.2\,\%), potentially due to architectural simplifications or instruction tuning compromises aimed at latency and efficiency. The superior performance of \texttt{Gemini-Flash-2.5} over \texttt{Flash-2.0} and \texttt{Flash‑Lite-2.0} \citep{comanici2025gemini25pushingfrontier} reflects incremental training improvements; and \texttt{Claude-3.7-Sonnet} (91.2\%) outperforming \texttt{Claude-3.5-Haiku} \citep{anthropic-claude3.5-haiku} (71.6\%) aligns with Anthropic’s published capability tiers.

To further investigate the potential of instruction-tuned open models, we fine-tuned \texttt{LLaMA3.3-70B-Instruct} and \texttt{Qwen3-32B-Instruct} by applying LoRA \citep{hu2022lora} on the attention layers, leveraging only 10\,\% of the training data (8{,}000 samples) for a sequence classification objective. Despite its poor zero-shot performance on strict accuracy (11.23\,\%), the fine-tuned \texttt{LLaMA3.3-70B-Instruct} achieved an accuracy of 86.82\,\%, while \texttt{Qwen3-32B} reached 85.64\,\%—both surpassing \texttt{DeepSeek-R1} (82.51\,\%) and \texttt{DeepSeek-V3} (82.41\,\%), the strongest open-source models in our zero-shot evaluation. This result highlights the latent capability of instruction-tuned LLMs and demonstrates that even lightweight, resource-efficient fine-tuning can substantially improve both task performance and output format adherence. Full fine-tuning details are provided in Appendix~\ref{sec:llama3-finetuning}.

\begin{table}[t]
\small
\centering
\caption{Accuracy of closed-source and open-source models on the test split of the \textsc{PerCoR} dataset.}
\label{tab:percor-eval}
\begin{adjustbox}{width=\columnwidth}
    \begin{tabular}{c l|cc}
\toprule
\textbf{Group} & \textbf{Model} & \textbf{Str Acc} & \textbf{PP Acc} \\
\midrule
\multirow{12}{*}{\rotatebox[origin=c]{90}{\textbf{Closed-Source}}} 
& GPT-4o-mini        &       75.98         &       75.98         
\\
& GPT-4o      &       78.32       &         86.65 
\\
& GPT-4.1-nano       &     54.94           &  54.94  \\
& GPT-4.1-mini       &    77.12            &       77.12         
\\
& GPT-4.1     &       88.39       &           88.39
\\
& OpenAI o3 & \textbf{92.18} & \textbf{92.18}
\\
& OpenAI o4-mini & 85.51 & 85.51
\\
& Gemini 2.0 Flash-Lite & 81.43 & 81.43
\\
& Gemini 2.0 Flash & 86.38 & 86.38
\\
& Gemini 2.5 Flash & 87.17 & 87.14
\\
& Claude 3.5 Haiku   &    71.60            &        71.60
\\
& Claude 3.7 Sonnet  &     \underline{91.17}           &       \underline{91.17}         \\
\midrule
\midrule
\multirow{20}{*}{\rotatebox[origin=c]{90}{\textbf{Open-Source}}} 
& Gemma 3n-E4B-it        &   59.15              &         59.15       \\
& Gemma 3-1B-it         &   25.99              &         25.99       \\
& Gemma 3-4B-it            &   48.32             &       48.32         \\
& Gemma 3-12B-it           &       70.94         &      70.94          \\
& Gemma 3-27B-it           &     76.28           &       76.28         \\
& Mistral 7B Instruct v0.3  &   30.11             &       30.15         \\
& Mistral Small 3.1 24B Instruct     &    68.94            &        68.94        \\
& LLaMA 3.2 1B Instruct    &      0.79          &        24.12        \\
& LLaMA 3.2 3B Instruct    &     25.17           &        25.21        \\
& LLaMA 3.3 70B Instruct    &     11.23           &        79.56        \\
& Aya Expanse 32B          &      5.85          &       63.27         \\
& Command R-v01          &      60.0         &     60.0          \\
& Command A          &      \underline{79.81}         &     79.84          \\
& Qwen 3-4B          &      50.33        &     50.33          \\
& Qwen 3-8B          &      54.37        &     54.37          \\
& Qwen 3-14B          &      69.58        &     69.58          \\
& Qwen 3-30B-A3B          &      68.80        &     68.80          \\
& Qwen 3-32B          &      76.54         &     76.54          \\
& DeepSeek-V3      &        51.15        &   \underline{82.41}              \\
& DeepSeek-R1      &        \textbf{82.51}        &   \textbf{82.51}              \\
\bottomrule
\end{tabular}
\end{adjustbox}
\end{table}

\subsection{Qualitative Error Analysis}
Despite strong overall performance, even top models occasionally fail on examples requiring subtle syntactic, temporal, or discourse-level reasoning. Appendix Figures~\ref{fig:earth-mars-mistake}--\ref{fig:phone-mistake} present four illustrative failure cases in which the model-selected continuation is semantically implausible, grammatically incompatible, or violates the discourse relation implied by the conjunction. Each example includes a brief explanation of why the chosen option is incorrect and why the gold continuation better satisfies the coherence constraints. These qualitative observations reveal failure patterns that are not apparent from aggregate accuracy alone and highlight the value of \textsc{PerCoR} for probing nuanced reasoning and alignment behaviours in large language models.

\subsection{Transferability to Persian QA}
Prior work such as \textsc{Social IQA} \citep{sap-etal-2019-social} and \textsc{WinoGrande} \citep{SakaguchiWinnogrande} has demonstrated transferability by fine-tuning models on a commonsense benchmark and evaluating on other reasoning tasks (e.g., \textsc{COPA} \citep{Gordon2011ChoiceOP} and \textsc{KnowRef} \citep{emami-etal-2019-knowref}). Such evaluation is not currently feasible in Persian, as \textsc{PerCoR} is the first structured commonsense reasoning benchmark in the language. We hope \textsc{PerCoR} will serve as a foundation that enables similar cross-task studies in the future. To provide an initial signal of transferability under current resource constraints, we evaluate on \textsc{PQUAD}~\citep{Darvishi_2023}, a Persian extractive QA benchmark in which many queries require implicit inference and bridging context. We fine-tune \texttt{Qwen3-32B} using LoRA with a causal language modelling objective on \textsc{PerCoR}, then zero-shot evaluate both the base and fine-tuned models on the \textsc{PQUAD} test set (964 examples) using Exact Match (EM), ROUGE-L \citep{lin-2004-rouge}, and BLEU \citep{papineni-etal-2002-bleu}.

Fine-tuning on \textsc{PerCoR} yields a small improvement in EM (25.21→26.56), but substantially increases ROUGE-L (19.27→29.69, +10.4) and BLEU (16.00→23.98, +7.9). These gains suggest that the discourse-level coherence learned from \textsc{PerCoR} transfers to extractive QA settings where models must integrate contextual cues with implicit knowledge. We view this as an encouraging early signal of transferability and include it to underscore \textsc{PerCoR}’s utility beyond its primary role as an evaluation benchmark.

In summary: \textbf{(i)} while \textsc{PerCoR} is approachable for top proprietary systems, it remains challenging for open-source models—only two models exceed 90\,\% accuracy, whereas the strongest open-source model still trails by approximately ten points; \textbf{(ii)} post-processing is often essential for revealing latent reasoning ability in models that express answers in free-form text rather than the required label format; \textbf{(iii)} reasoning-oriented alignment matters—\texttt{OpenAI-o3} leads all systems, and \texttt{DeepSeek-R1} surpasses \texttt{DeepSeek-V3} in strict accuracy due to stronger format adherence; \textbf{(iv)} lightweight task-specific fine-tuning remains highly effective, with small adaptations outperforming the strongest open-source zero-shot baselines; \textbf{(v)} even top-performing models exhibit systematic mistakes on examples requiring subtle syntactic, temporal, or discourse-level reasoning, underscoring \textsc{PerCoR}'s diagnostic value; and \textbf{(vi)} the discourse-level coherence captured by \textsc{PerCoR} shows early signs of transferability to other Persian tasks, as reflected in improvements on \textsc{PQUAD}.

%% file: sections/conclusion.tex
\section{Conclusion}

We introduced \textbf{\textsc{PerCoR}}, a 106K-example benchmark that fills a major evaluation gap for commonsense reasoning in Persian. Our conjunction-based extraction strategy yields natural sentence–completion pairs from diverse prose, and DRESS-AF selects challenging, human-written distractors while avoiding stylistic artifacts associated with LLM-generated options. Evaluating 32 models reveals a persistent ten-point gap between the strongest open-source and proprietary systems, while qualitative analysis exposes systematic failures on examples requiring subtle syntactic, temporal, or discourse-level reasoning. Together with early signs of transferability to downstream Persian tasks, these results show that \textsc{PerCoR} provides a robust benchmark, a fine-grained diagnostic tool, and a valuable resource for transfer learning in low-resource settings.

Future work will (i) extend our language-agnostic pipeline to other languages by adapting conjunction lists and applying DRESS-AF, and (ii) conduct expert-based human evaluation to establish a high-quality gold standard for ambiguous cases. We believe \textsc{PerCoR} will catalyse research on multilingual commonsense reasoning and foster the development of more robust, culturally-aware language models.

%% file: sections/limitation.tex
\paragraph{Annotation}
As noted previously (see Section \ref{sec:experiments}), our annotations were conducted by human annotators rather than human experts. While this approach is sufficient for broad evaluations, relying on expert annotators would likely yield more accurate and reliable assessments, particularly for complex or ambiguous cases. Moreover, we could have annotated a larger portion of the dataset to obtain a more robust and reliable estimate of human accuracy. Additionally, we could have adopted a standard annotation strategy similar to the one used in HellaSwag \citep{zellers-etal-2019-hellaswag}, which involves multiple rounds of human validation and a larger set of possible answers to choose from. However, this approach requires substantially more human effort and coordination, making it more resource-intensive.

\paragraph{Multilingual}
Given that the proposed method is largely language-agnostic, we could have extended the algorithm to other languages to construct a multilingual commonsense reasoning dataset. This would have involved creating lists of conjunctions in each target language for the sentence-completion step, followed by applying the DRESS-AF algorithm accordingly.

%% file: sections/ethics.tex
\paragraph{License}

In accordance with OpenAI’s Terms of Use, “as between you and OpenAI... you (a) retain your ownership rights in Input and (b) own the Output. We hereby assign to you all our right, title, and interest, if any, in and to Output”\footnote{\url{https://openai.com/policies/row-terms-of-use/}}.\

Google Gemini’s terms distinguish between paid vs unpaid usage: under paid/enterprise tiers, Google does not use submitted prompts or outputs to train its models and customers retain ownership of both input and output\footnote{\url{https://ai.google.dev/gemini-api/terms}}. Under unpaid or free tiers, Google may use content for product improvements, and retention policies differ.\

Anthropic’s Claude Terms grant users ownership of all generated outputs: “subject to your compliance with our Terms, we assign to you all of our right, title, and interest—if any—in Outputs”\footnote{\url{https://terms.law/2024/08/24/who-owns-claudes-outputs-and-how-can-they-be-used/}}.\

Based on these platform policies, we acknowledge that—under the Terms of Use for OpenAI, Google Gemini (paid/enterprise tiers), and Anthropic Claude—users retain ownership of both prompts (inputs) and generated outputs, and that the AI‑produced text used in this research was obtained and employed ethically within those licensing frameworks.

Furthermore, we confirm that the outputs generated from the model were not used to train or develop models that compete with These Models. All content and model-generated assistance were applied solely for academic and illustrative purposes in the context of this research.

To generate the PerCoR dataset, we utilized textual data extracted from over 40 publicly accessible websites. All selected sources were openly available and did not impose restrictions that would preclude academic or non-commercial use. 

\paragraph{Harmful content}
To curate our dataset, we selected sources with minimal sexual content and hate speech to maintain ethical standards. However, due to the complexities of open-domain language and commonsense reasoning tasks, we cannot guarantee the absence of social biases. As noted in prior work \citep{sakai-etal-2024-mcsqa, rajani-etal-2019-explain, sap-etal-2020-social}, it remains challenging to determine when content that reflects commonsense also constitutes social bias.

%% file: sections/appendix.tex
\section{Dataset Creation}
\label{sec:appendix-ds-creation}

\subsection{Sentence-Completion Filtering}
\label{sec:appendix-completion}

\begin{figure}[h]
    \centering
    \includegraphics[width=\columnwidth]{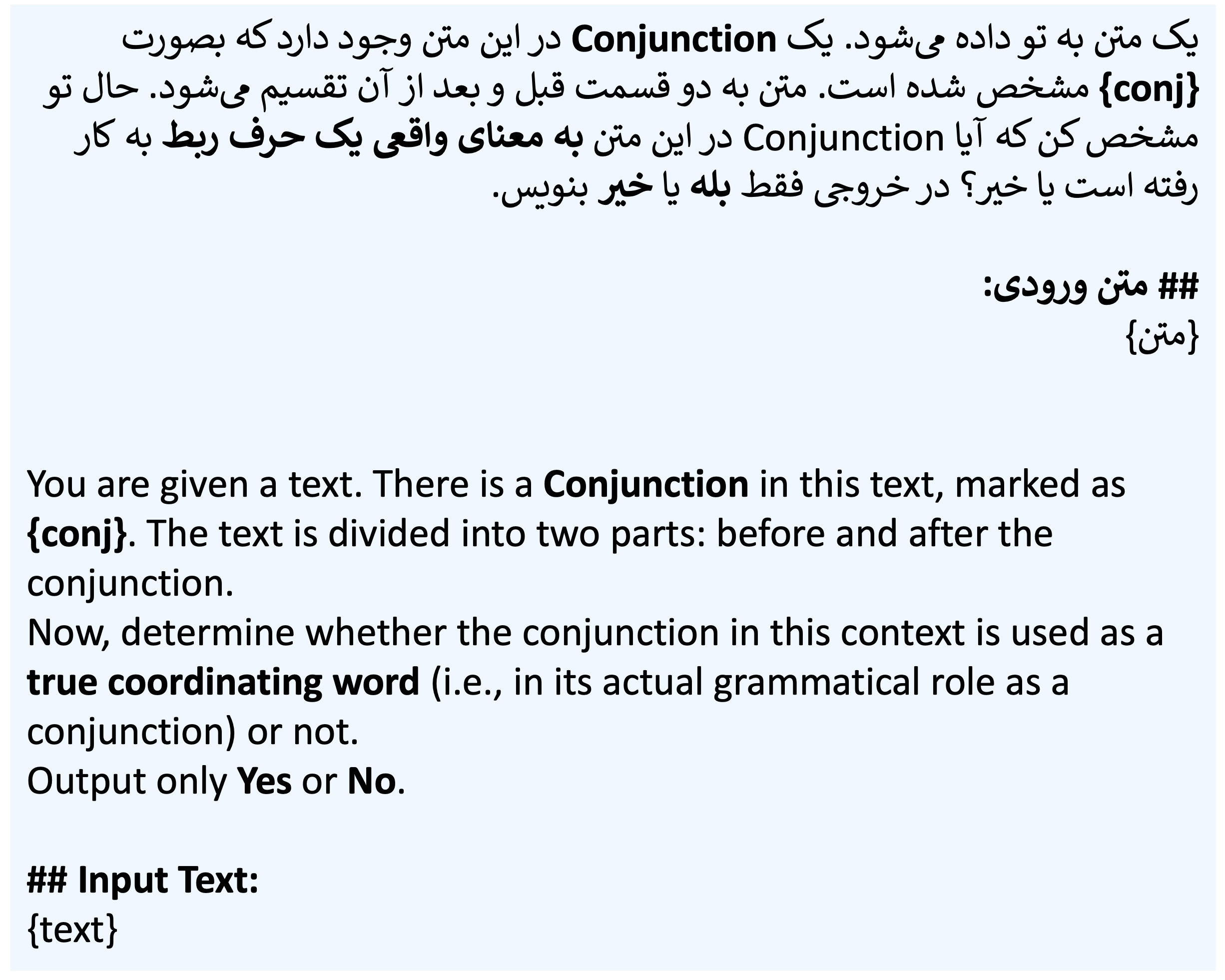}
    \caption{The prompt that we used for GPT4o-mini to detect the ambiguity of conjunctions.}
    \label{fig:conj_promp}
\end{figure}

\begin{figure}[h]
    \centering
    \includegraphics[width=\columnwidth]{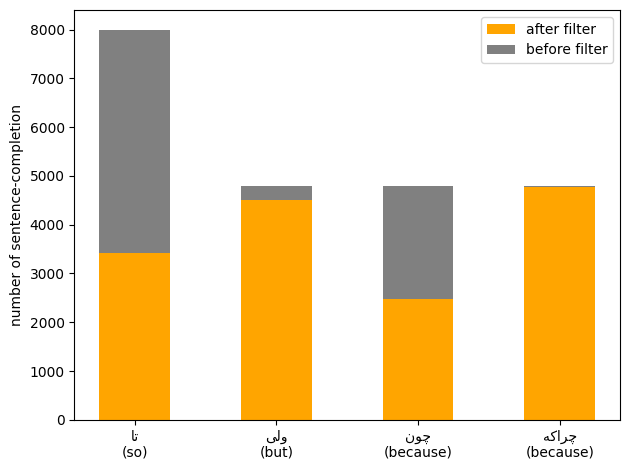}
    \caption{The number of data instances with ambiguous conjunctions before and after filtering with the GPT4o-mini model.}
    \label{fig:cm_filter}
\end{figure}

Figure \ref{fig:cm_filter} presents a list of Persian conjunctions that exhibit semantic ambiguity. To address this, we employ GPT4o-mini using a binary classification prompt (shown in Figure~\ref{fig:conj_promp}) to filter out sentences in which the conjunction is used with a meaning other than the intended one. The final proportion of retained data for each connective after this filtering step is also reported in Figure \ref{fig:cm_filter}.

\begin{figure*}[ht]
    \centering
    \includegraphics[width=\textwidth]{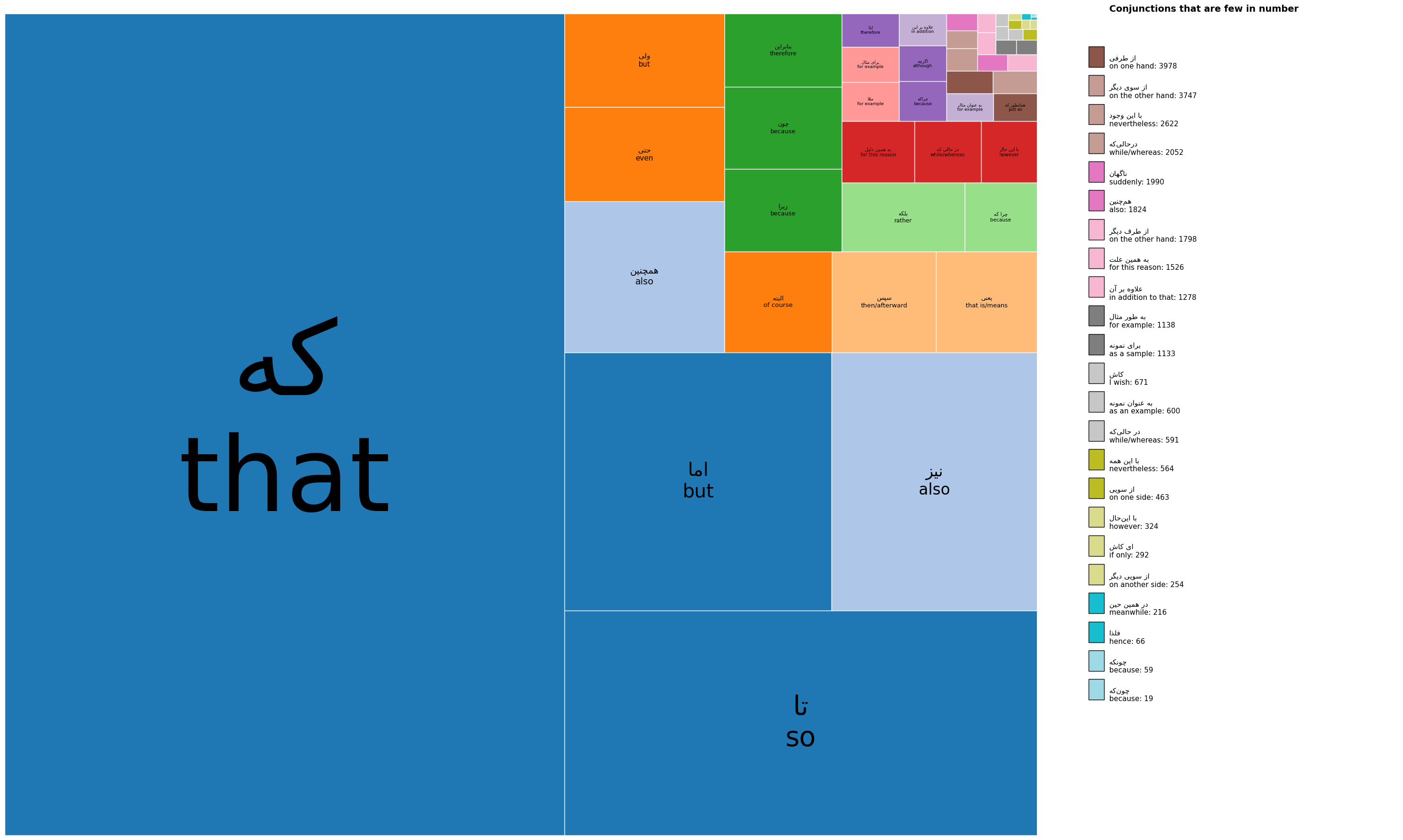}
    \caption{Treemap visualisation of conjunction words used to generate sentence–completion pairs. The area of each block corresponds to the conjunction's frequency in the dataset. Less frequently used conjunctions are shown in the adjacent panel for completeness.}
    \label{fig:conjunction_list}
\end{figure*}

\begin{figure}[h]
    \centering
    \includegraphics[width=\columnwidth]{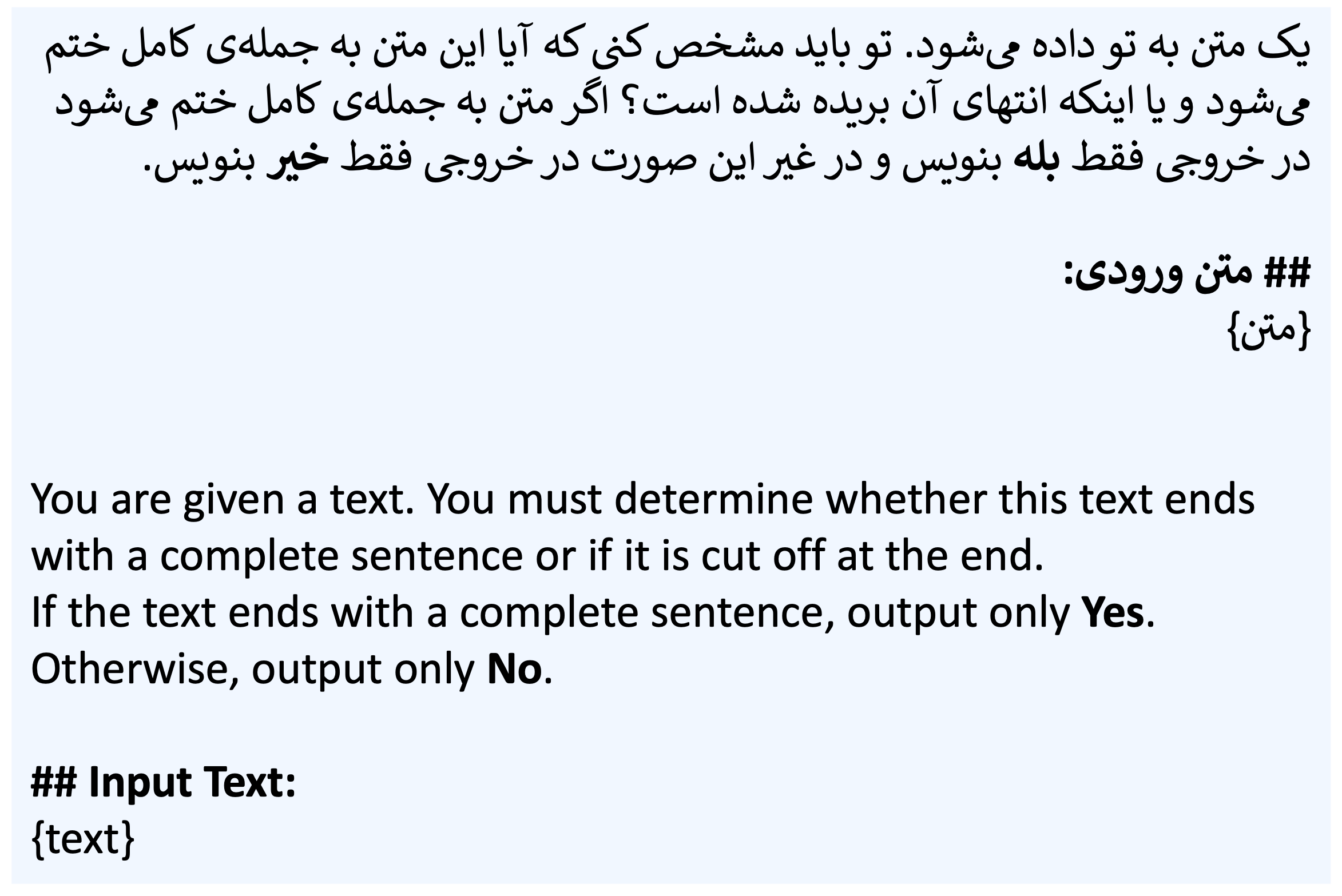}
    \caption{The prompt that is leveraged by GPT4o-mini to filter out the incomplete pairs.}
    \label{fig:comp_prompt}
\end{figure}

Furthermore, we perform an additional pass over the entire dataset using a second prompt (shown in Figure~\ref{fig:comp_prompt}) to identify structurally incomplete sentence completions. At this stage, 12,117 out of 135,912 instances are flagged by the model as incomplete and removed from the dataset accordingly.

\section{Model Evaluation}
\label{sec:appendix-model-eval}

\subsection{Model Configurations}
\label{sec:appendix-model-config}

We conducted all evaluations using the \texttt{vLLM} inference engine for efficient serving of open-source models. Each model was run on a single NVIDIA A100 80GB GPU, except for LLaMA 3.2 70B Instruct, which required two A100 80GB GPUs due to its size. For DeepSeek variants, we used their official API endpoints, as the open-source checkpoints were not served locally.

\subsection{Model Behaviour by Input Length}
\label{sec:appendix-model-len}

\begin{figure}[h]
    \centering
    \includegraphics[width=\columnwidth]{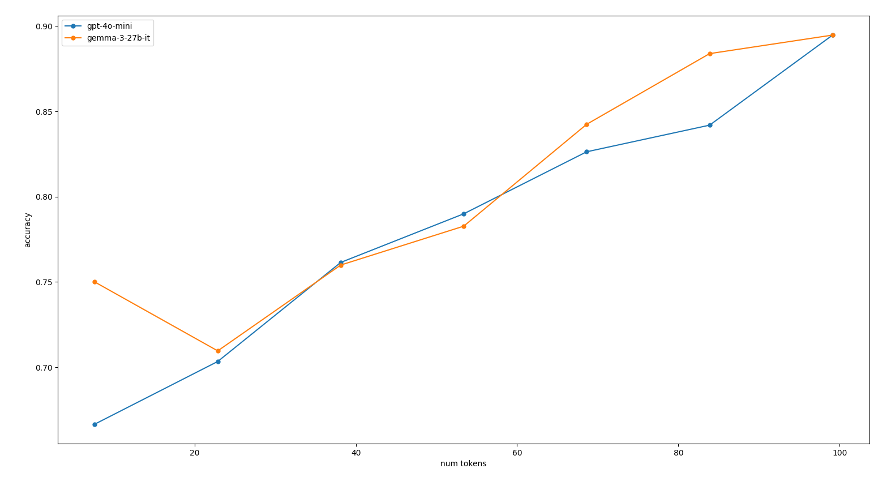}
    \caption{Accuracy of GPT-4o-mini and Gemma 3-27B-it as a function of input length. Longer prefixes tend to improve accuracy by offering more contextual information.}
    \label{fig:acc-based}
\end{figure}

Owing to our conjunction-based segmentation strategy, \textsc{PerCoR} samples exhibit a broad range of input lengths. To assess how input length influences model performance, we analyse the correlation between sentence length and model accuracy. As shown in Figure~\ref{fig:acc-based}, both GPT-4o-mini and Gemma 3-27B-it exhibit improved accuracy as the number of input tokens increases.

This trend suggests that longer sentence prefixes provide more contextual cues, enabling models to more reliably identify the correct completion. The result highlights a natural advantage for models when reasoning over richer, more informative contexts—an important factor to consider when designing evaluation datasets for commonsense reasoning.

\begin{figure}[htbp]
    \centering
    \includegraphics[width=\columnwidth]{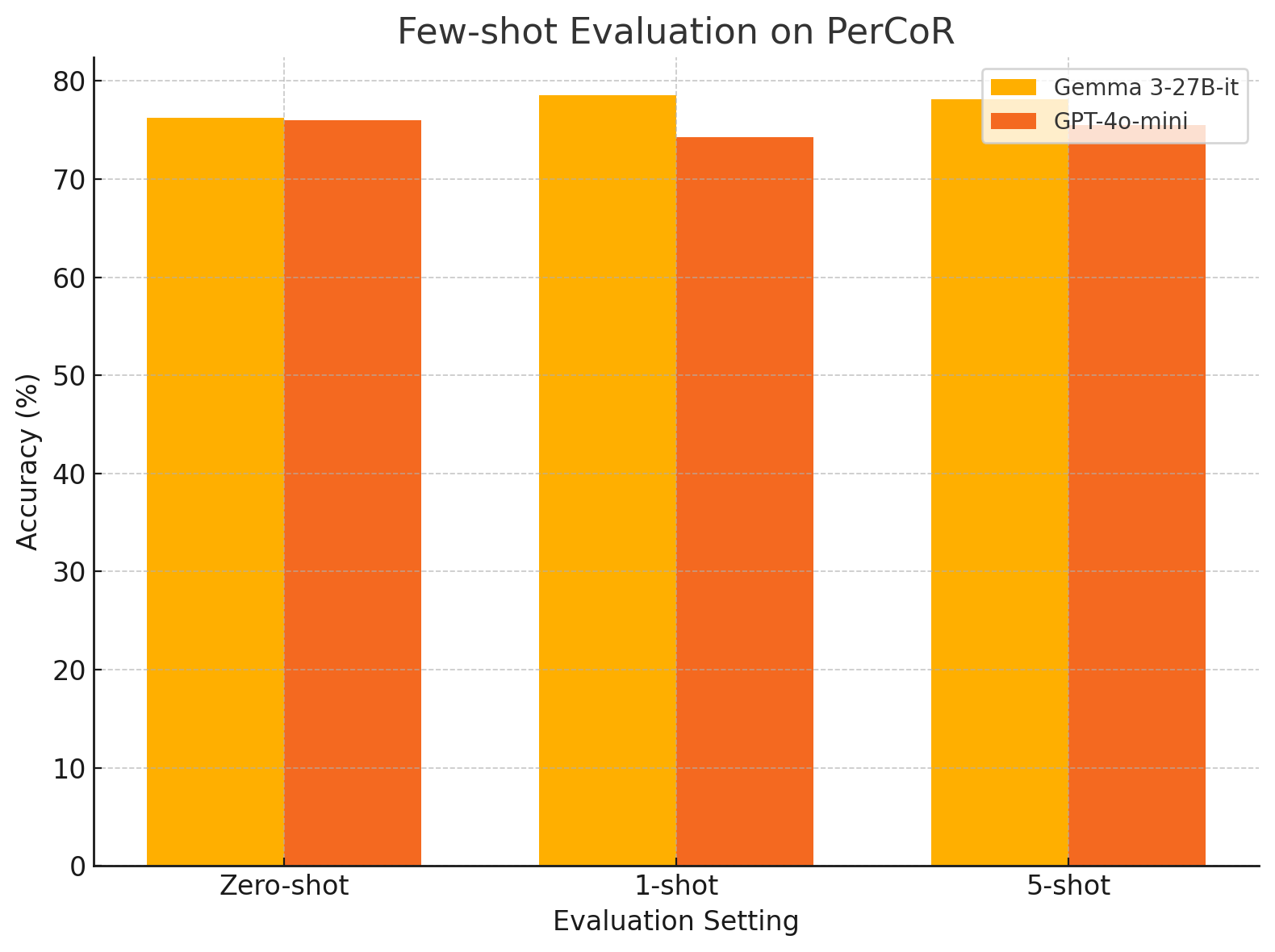}
    \caption{Accuracy of \texttt{GPT-4o-mini} and \texttt{Gemma3-27B-it} on the \textsc{PerCoR} dataset under zero-shot, 1-shot, and 5-shot settings.}
    \label{fig:few-shot}
\end{figure}

\subsection{Few-Shot Performance}
\label{sec:appendix-fewshot}

To assess model sensitivity to minimal supervision, we conducted 1-shot and 5-shot evaluations on the PerCoR dataset using \textit{GPT-4o-mini} and \textit{Gemma 3-27B-it}. As shown in Figure~\ref{fig:few-shot}, Gemma benefits modestly from few-shot prompting, improving from 76.28\% (zero-shot) to 78.51\% (1-shot) and 78.17\% (5-shot). In contrast, GPT-4o-mini exhibits marginal or inconsistent gains, with accuracy fluctuating around its zero-shot baseline of 75.98\%. These results highlight the robustness of Gemma to few-shot prompting and suggest that further gains may require stronger prompt design or fine-tuning.

\subsection{Human Evaluation}
\label{sec:appendix-humaneval}
We used Label Studio \cite{labelstudio} to evaluate the accuracy on the test split of the dataset. Each sample was annotated independently by three human annotators. In cases where at least two annotators agreed on the same label, their consensus was taken as the final label, which was then compared with the provided label. If all three annotators disagreed, the sample was considered incorrectly labelled. Importantly, annotators worked independently and were not aware of each other's selections.

\subsection{Model Fine-tuning on the Dataset}
\label{sec:llama3-finetuning}

To evaluate the impact of fine-tuning on \textsc{PerCoR}, we selected two instruction-tuned open-source models: \texttt{LLaMA-3.3-70B-Instruct} and \texttt{Qwen3-32B}. The former was quantised to 4-bit precision, while the latter was trained using \texttt{bfloat16}. We used a per-GPU batch size of 8 for \texttt{LLaMA} and 4 for \texttt{Qwen3}. Training was conducted for 2 epochs using 8$\times$A100 80GB GPUs with DeepSpeed \citep{deepspeed2020} for distributed optimisation. We used HuggingFace \citep{wolf2020huggingfacestransformersstateoftheartnatural} for training the models.

Both models were fine-tuned using a Cosine learning rate scheduler with an initial learning rate of 5e--5 and a warmup ratio of 0.03. LoRA \citep{hu2022lora} was applied to the \texttt{q}, \texttt{k}, \texttt{v}, and \texttt{o} projection matrices within the attention layers, with hyperparameters $r{=}4$ and $\alpha{=}8$.

\begin{figure}[h]
    \centering
    \includegraphics[width=\columnwidth]{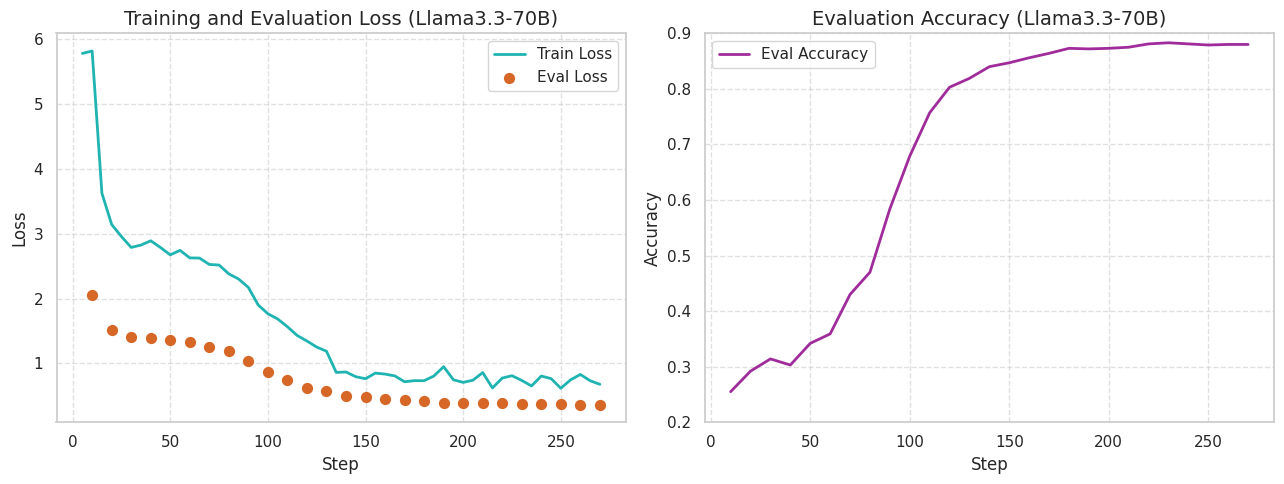}
    \caption{Training/evaluation loss and evaluation accuracy during fine-tuning of \texttt{LLaMA-3.3-70B-Instruct} on \textsc{PerCoR}.}
    \label{fig:llama_ft}
\end{figure}

\begin{figure}[h]
    \centering
    \includegraphics[width=\columnwidth]{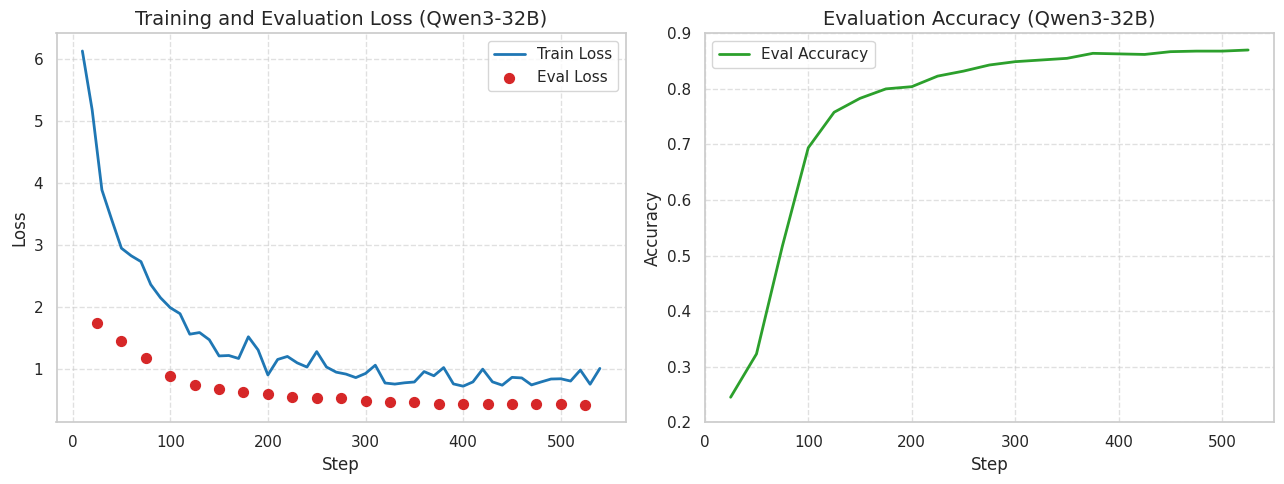}
    \caption{Training/evaluation loss and evaluation accuracy during fine-tuning of \texttt{Qwen3-32B-Instruct} on \textsc{PerCoR}.}
    \label{fig:qwen_ft}
\end{figure}

Figures~\ref{fig:llama_ft} and~\ref{fig:qwen_ft} show the training loss, evaluation loss, and evaluation accuracy over the course of training for both models. Training took approximately 2.5 hours for \texttt{LLaMA3.3} and around 1 hour for \texttt{Qwen3}.

\begin{figure*}[h]
    \centering
    \includegraphics[width=\textwidth]{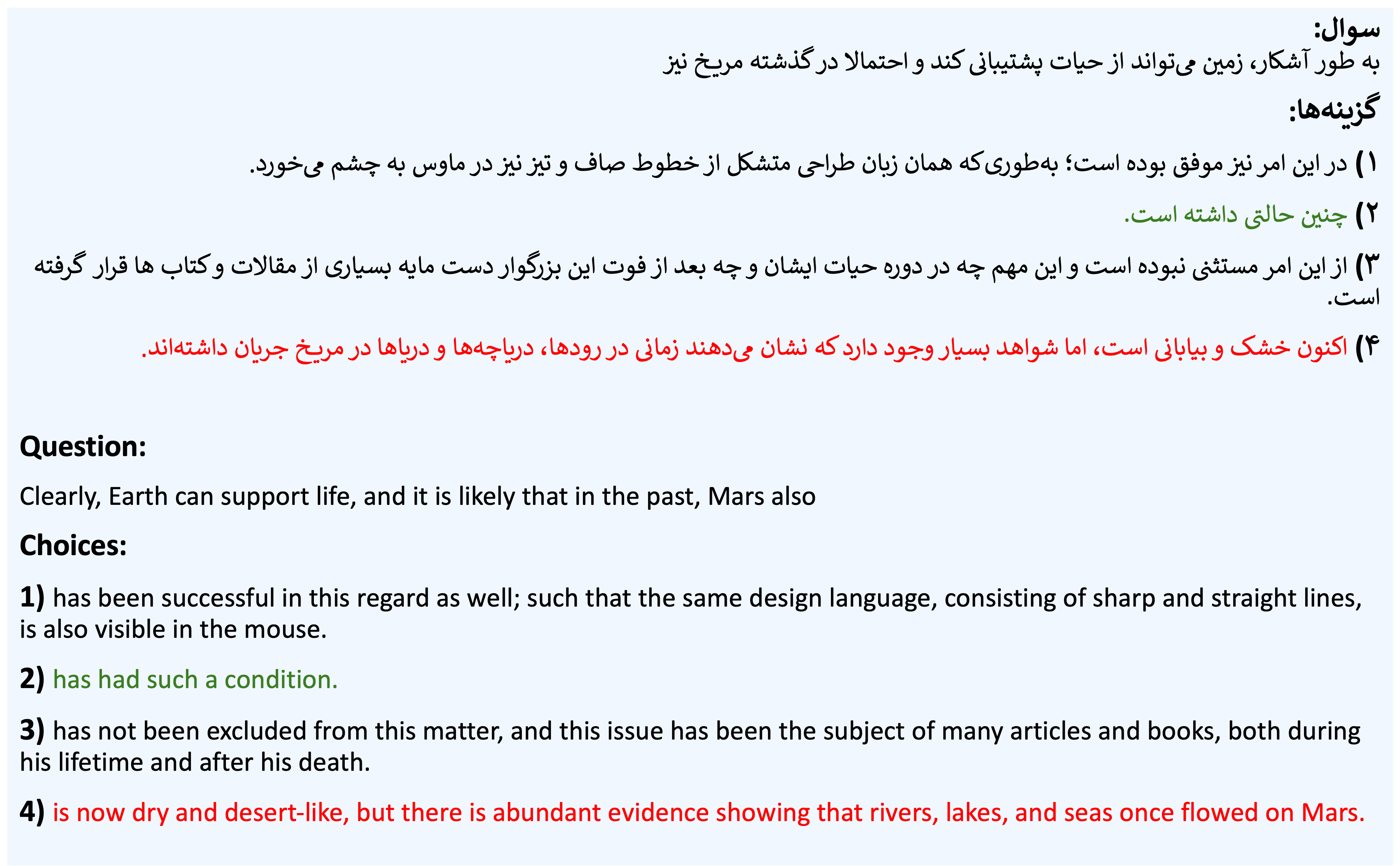}
    \caption{An example from the PerCoR dataset where both Claude 3.7 Sonnet and GPT-4.1 incorrectly selected Option 4 (highlighted in red). While Option 4 contains true statements about the current and past state of Mars, it fails to form a coherent continuation when appended to the prompt. The question sets up a comparison referring specifically to Mars’s past, expecting a grammatically and temporally consistent continuation. Option 2 correctly completes the sentence with a minimal and coherent reference to Mars’s possible past habitability.}
    \label{fig:earth-mars-mistake}
\end{figure*}

\begin{figure*}[h]
    \centering
    \includegraphics[width=\textwidth]{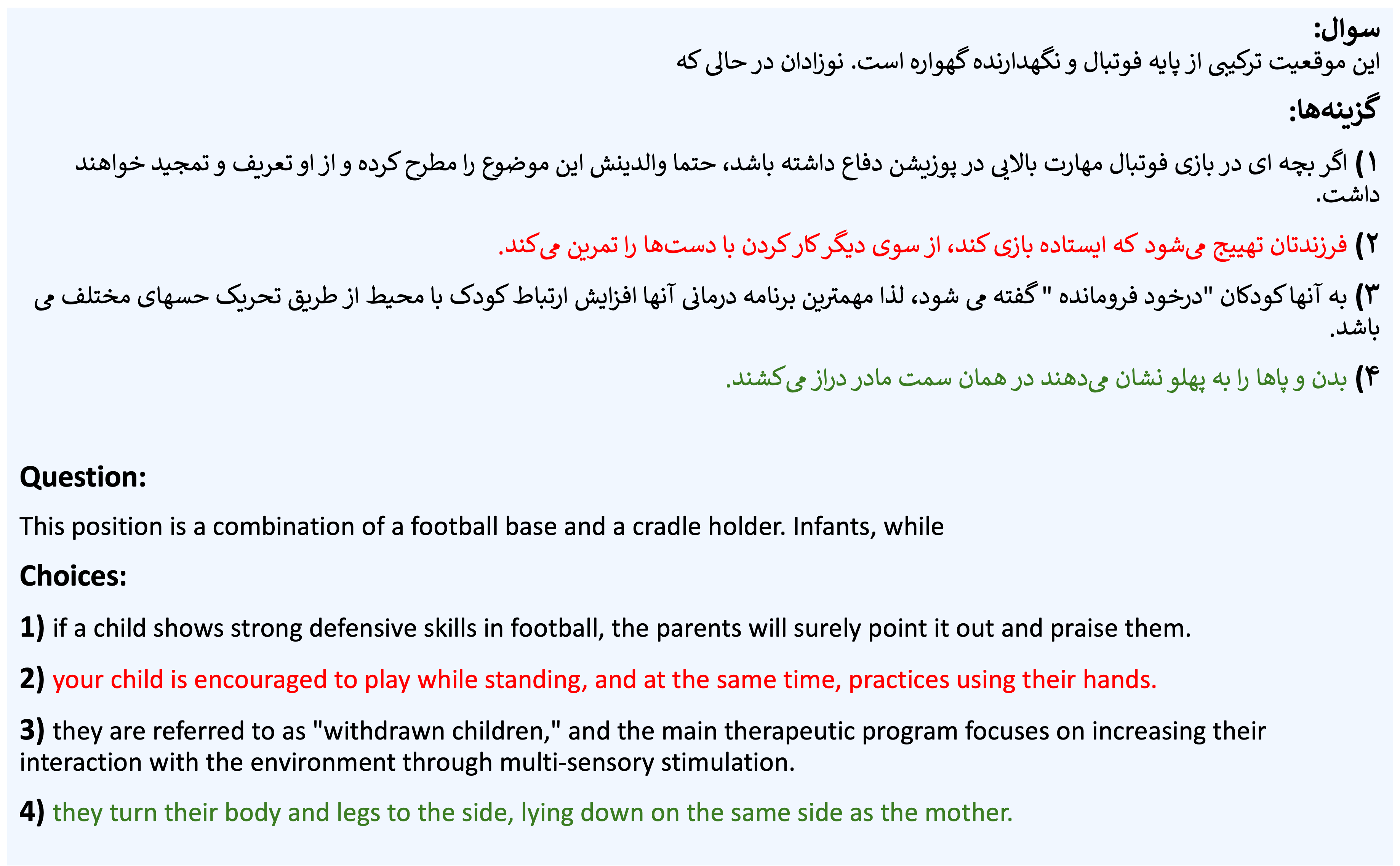}
    \caption{An example from the PerCoR dataset where both Claude 3.7 Sonnet and GPT-4.1 incorrectly selected Option 2 (highlighted in red). Appending Option 2 to the prompt results in a grammatically broken and incoherent sentence: “Infants, while your child is encouraged to play while standing...” — which abruptly shifts subject and verb, making no syntactic or semantic sense. The phrase “Infants, while...” requires a continuation that describes a physical or observational state of the infant. Only Option 4 satisfies this expectation with a coherent and contextually appropriate description of the infant’s posture.}

    \label{fig:infant-mistake}
\end{figure*}

\begin{figure*}[h]
    \centering
    \includegraphics[width=\textwidth]{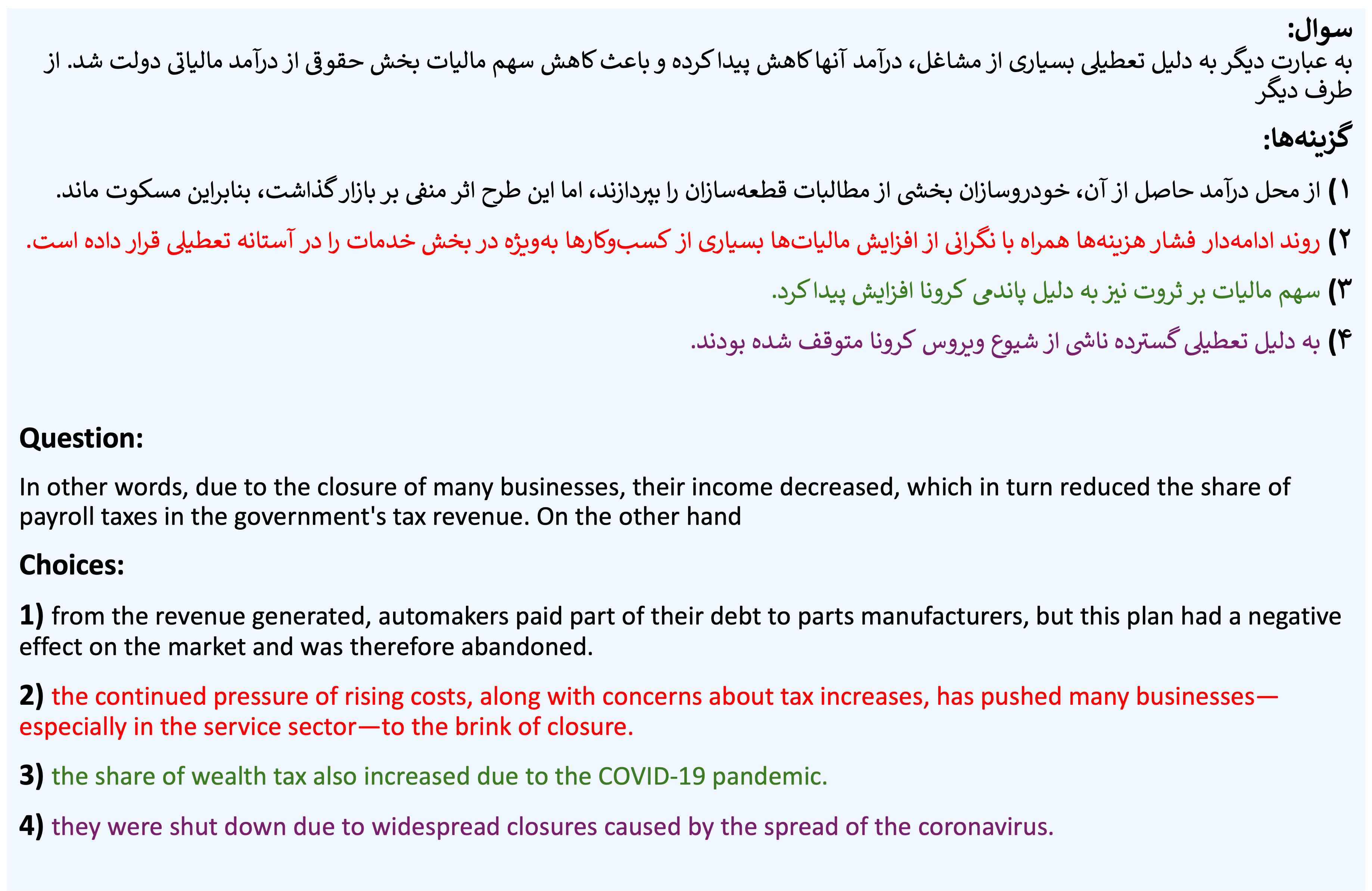}
    \caption{An example from the PerCoR dataset where Claude 3.7 Sonnet and GPT-4.1 both incorrectly selected Options 2 and 4, respectively. The sentence discusses how business closures led to a decline in payroll tax contributions, and the phrase “On the other hand...” introduces a contrasting development that should remain within the domain of \textbf{tax revenue}. While Option 2 is contextually plausible—highlighting economic stress—it shifts the focus away from taxation. Option 4 is even less relevant, as it redundantly repeats the cause already stated in the prompt (business closures due to COVID-19). In contrast, Option 3 presents a coherent and contrastive continuation: despite payroll tax revenue declining, the share of wealth tax increased during the pandemic. This makes Option 3 the most topically and logically aligned completion.}

    \label{fig:tax-mistake}
\end{figure*}

\begin{figure*}[h]
    \centering
    \includegraphics[width=\textwidth]{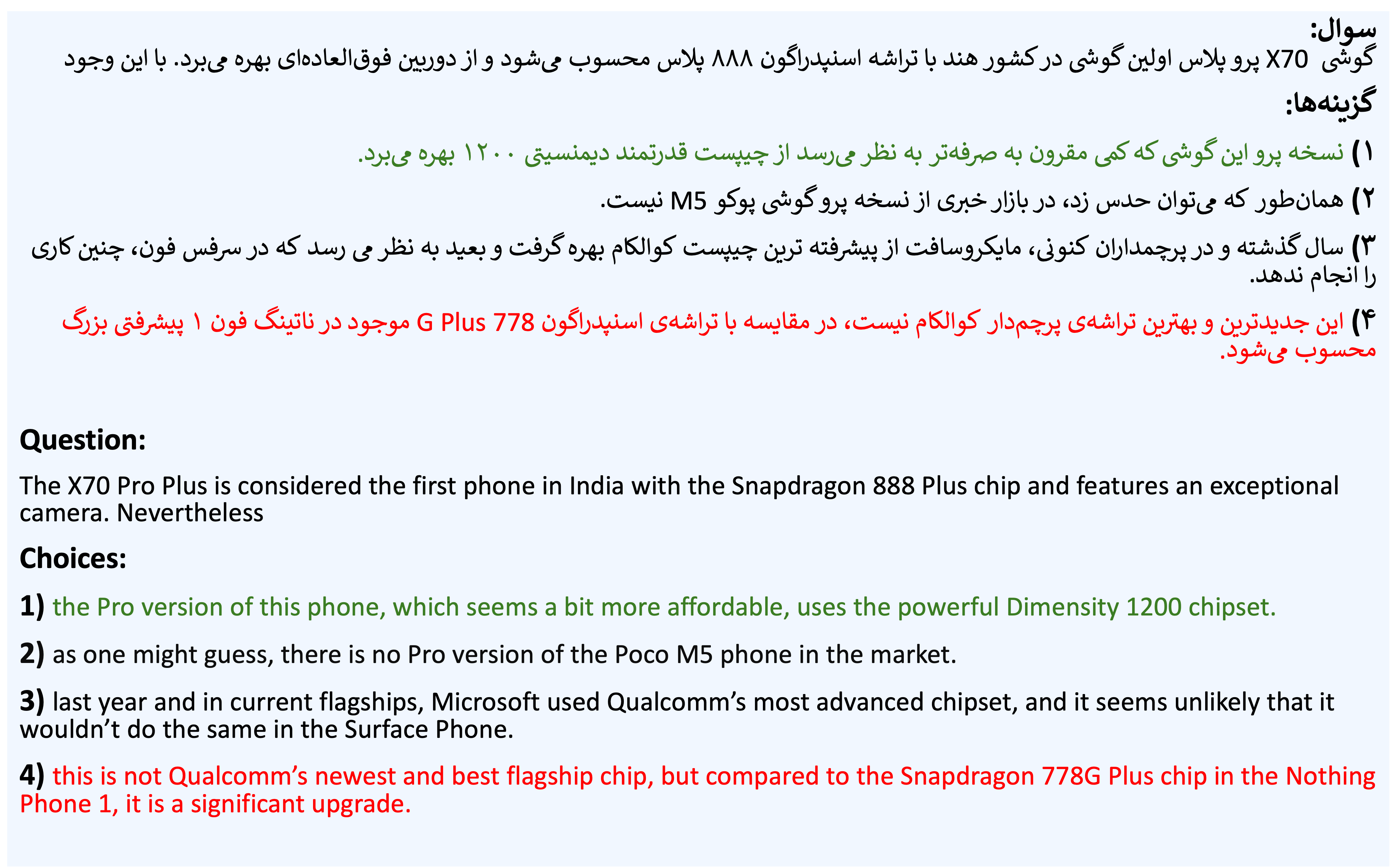}
    \caption{An example from the PerCoR dataset where Claude 3.7 Sonnet incorrectly selected Option 4 as the answer. While Option 4 provides a comparison between chipsets, it fails to directly continue the original sentence, which is about the X70 Pro Plus smartphone. The phrase “Nevertheless...” sets up a contrast or qualification specifically about the phone mentioned. Option 1 correctly continues this contrast by discussing the Pro variant of the same phone and its different chipset—maintaining topical and grammatical coherence. In contrast, Option 4 shifts focus entirely to the chipset itself, breaking the discourse continuity and making it an incoherent continuation in context.}
    \label{fig:phone-mistake}
\end{figure*}